# Elements of Robot Morphology: Supporting Designers in Robot Form Exploration

Amy Koike
University of Wisconsin-Madison
Madison, USA
ekoike@wisc.edu

Serena Ge Guo
University of Wisconsin-Madison
Madison, USA
gguo28@wisc.edu

Xinning He
University of Wisconsin-Madison
Madison, USA
xinning.he@wisc.edu

Callie Y. Kim
University of Wisconsin-Madison
Madison, USA
cykim6@cs.wisc.edu

Dakota Sullivan
University of Wisconsin-Madison
Madison, USA
dsullivan8@wisc.edu

Bilge Mutlu
University of Wisconsin-Madison
Madison, USA
bilge@cs.wisc.edu

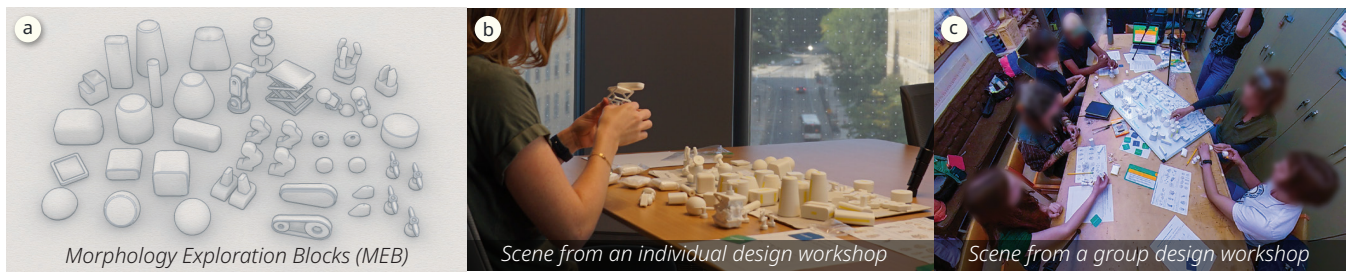

Figure 1: We propose `Elements of Robot Morphology`—a framework of the basic elements that make up robot morphology. To bring this framework into practice, (a) we developed `Morphology Exploration Blocks (MEB)`—a tangible toolkit designed to facilitate hands-on form exploration. The design of these blocks was informed by an extensive analysis of existing robots. We evaluated the framework and MEB through both (b) individual design workshops and (c) group design workshops.

## Abstract

Robot morphology–the form, shape, and structure of robots–is a key design space in human–robot interaction (HRI), shaping how robots function, express themselves, and interact with people. Yet, despite its importance, little is known about how design frameworks can guide systematic form exploration. To address this gap, we introduce `Elements of Robot Morphology`, a framework that identifies five fundamental elements: perception, articulation, end effectors, locomotion, and structure. Derived from an analysis of existing robots, the framework supports structured exploration of diverse robot forms. To operationalize the framework, we developed `Morphology Exploration Blocks (MEB)`, a set of tangible blocks that enable hands-on, collaborative experimentation with robot morphologies. We evaluate the framework and toolkit through a case study and design workshops, showing how they support analysis, ideation, reflection, and collaborative robot design.

## 1 Introduction

Robot morphology—the form, body shape, and structure of robots—constitutes a key design space in human-robot interaction (HRI), as it is intertwined with how robots function, express themselves, interact with humans, and are perceived by them. Several frameworks have been proposed to describe and analyze robot morphology, such as the Abot database [44] and the Metamorph framework [46]. Formats like URDF (Unified Robot Description Format) [66], used within the Robot Operating System (ROS) ecosystem, can also serve as morphology frameworks. These resources enable researchers, designers, and engineers to describe, analyze, and compare a robot's physical features and appearance.

Despite this emerging body of work on *analytical* frameworks focused on understanding and representing robot morphology, little is known about how design frameworks might guide form exploration and generation. The prevailing paradigm in robot design currently involves careful exploration of appropriate *metaphors* [13, 37] that may serve as a conceptual foundation for user interaction with the robot, such as the approach proposed by Alves-Oliveira et al. [3]. However, little work has been done on how designers might morphologically realize these conceptual designs. Furthermore, design efforts that may not follow appropriate metaphors might still need support in form exploration. Thus, there is a need for frameworks that can serve as *analytical* tools to understand existing designs as well as *generative* tools to explore novel forms.

To address this need, we propose `Elements of Robot Morphology`, a framework inspired by the elements of graphic design [67]. Just as graphic design elements, such as "line," "shape," and "color," serve as foundational building blocks, our framework provides fundamental morphological elements that designers can combine to *analyze* and *generate* robot forms. Based on an analysis of robots in the IEEE robot database [28], we identify five categories of elements: **perception**, **articulation**, **end effectors**, **locomotion**, and **structure**.

To operationalize this framework, we developed `Morphology Exploration Blocks (MEB)`, a toolkit that allows designers to interact with and explore robot morphology through physical building blocks. MEB is designed to foster hands-on experimentation and collaborative exploration of robot forms (Figure 1(a)). We evaluated the framework and toolkit through two studies: a case study, which demonstrated how the framework can serve as an *analytical* tool, and design workshops, where designers used MEB to *generate and explore* robot designs. We conducted both individual and group workshops, observing how participants used the toolkit for individual ideation and collaboration. Our contributions are as follows: (1) **Framework:** `Elements of Robot Morphology`, a conceptual structure for describing robot form and *analytical* and *generative* processes to utilize it for robot design; (2) **Artifact:** `Morphology Exploration Blocks (MEB)`, a set of building blocks to make the framework operational; and (3) **Empirical:** A case study and workshops demonstrating the framework's application as both an *analytical* and *generative* tool for robot morphology design.

## 2 Related Work

### 2.1 Role of Robot Morphology

HRI research has extensively examined how a robot's appearance influences social perception, expectations, and user bias [*e.g.,* 10, 14, 22, 47]. Notably, Haring et al. [22] introduced form–function attribution bias (FFAB), showing that users often overestimate a robot's capabilities based on its form. Prior work has also compared perceptions of different robot morphologies and shapes [*e.g.,* 10, 18, 42], demonstrating effects on likability, perceived intelligence, and safety [9, 21]. Beyond categorical morphology, Hwang et al. [27] showed that overall robot shape alone can influence users' emotional responses and personality attribution.

In contrast, morphological computation research focuses on how a robot's physical form (*e.g.,* shape or material) contributes to its embodied capabilities, such as control, sensing, and perception, thereby reducing computational demands [*e.g.,* 39, 43].

### 2.2 Robot Morphology Framework

Beyond studies examining how robot morphology affects social perception and capability, prior work has focused on documenting and describing robot embodiment through databases, catalogs, and description schemes. Resources such as the Abot database [44], the IEEE Robot Guide [28], and the MUFaSAA dataset [14] compile existing robot designs and, in some cases, link embodiment to social and functional expectations. HRI research has also proposed taxonomies and conceptual frameworks to classify robot embodiment, ranging from broad categorical schemes [16, 38, 40, 51, 68] to detailed morphological descriptions [6, 44, 46]. For example, Fong et al. [16] introduced four embodiment categories–anthropomorphic, zoomorphic, caricatured, and functional–while Ringe et al. [46] proposed a framework for encoding fine-grained morphological features. Technical frameworks such as URDF [66] further support part-level description for simulation and control, but are not designed for expressive or exploratory design.

Table 1: Comparison of robot morphology frameworks and design tools relative to our proposed framework.

|  | Purpose | Abstraction | Tangible |
| --- | --- | --- | --- |
| URDF [66] | modeling/simulation | low | N/A |
| Metamorph [46] | description/analysis | mid–high | N/A |
| Metaphor-based [3, 36] | ideation/inspiration | very high | ✓ |
| **This Work** | analysis/exploration | mid | ✓ |

While these frameworks support the description, comparison, or modeling of robot bodies, few offer a design language that enables deliberate exploration of robot morphology (*i.e.,* balancing form, function, and interaction). Moreover, many of them rely on human or animal metaphors; in contrast, we aim to define morphological elements as functional, compositional building blocks (Table 1).

### 2.3 Robot Design Methods and Tools

HRI research has proposed diverse approaches to robot design, including participatory co-design [*e.g.,* 1, 5, 8, 34, 48, 52], Wizard-of-Oz prototyping [*e.g.,* 15, 45], and movement-centric design [*e.g.,* 23, 54]. While these approaches primarily offer conceptual or methodological guidance, toolkit-based studies [*e.g.,* 2, 4, 26, 53] enable hands-on exploration and function as both generative design aids and research instruments. For example, FLEXI [2] provides a customizable embodiment kit for rapidly prototyping social robots, while tangible scenography [30] uses physical props to support interaction design and role-play. Commercial platforms such as LEGO Mindstorms [56] and Fable [50] similarly lower barriers to robot construction, but primarily emphasize movement and behavior rather than deliberate exploration of robot morphology.

Although many existing design methods and tools emphasize behavior, motion, or functional prototyping, few treat form itself as a central design variable, particularly during early-stage exploration. Building on recent HRI research that expands the design space of robot form, such as work on soft robots [25, 29, 31], our work introduces a toolkit that enables HRI researchers and designers to directly engage with this broader landscape of robot morphology.

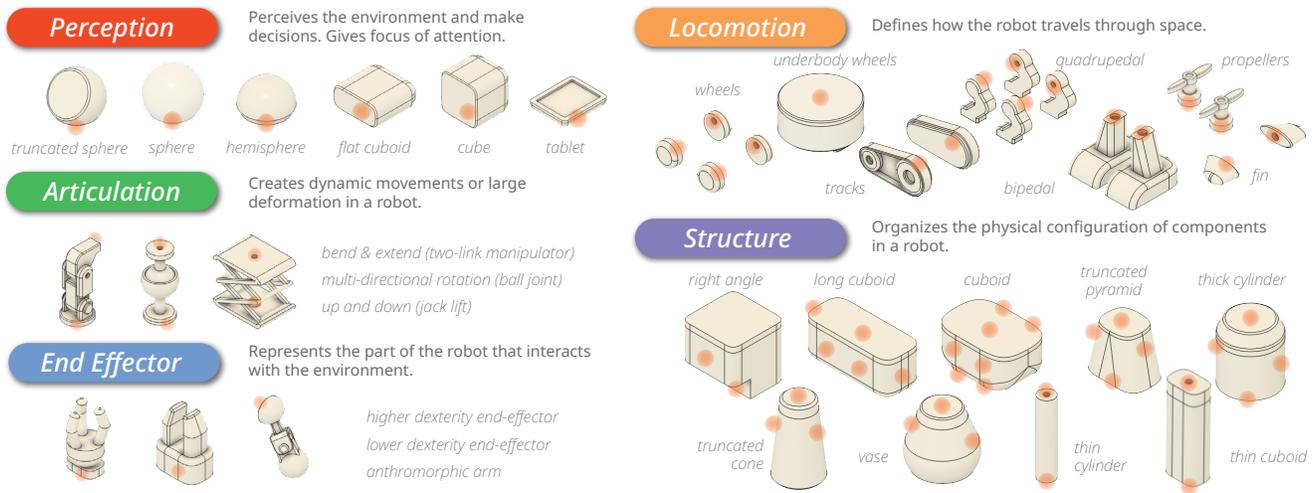

Figure 2: The five categories of Elements of Robot Morphology and the corresponding Morphology Exploration Blocks (MEB). The blocks abstract out and represent common components of existing robots, enabling designers to explore a wide variety of robot forms. Blocks were 3D-printed using PLA material and assembled with embedded magnets, marked by orange circles.

## 3 Elements of Robot Morphology

### 3.1 Framework Development

Our framework emerged from an iterative process aimed at answering the guiding question: "How can we guide robot form exploration and generation?" To address this question, we drew inspiration from *elements of graphic design* [67] and *elements of architecture* [33]. Just as graphic design elements–such as line, shape, color–and architectural elements–such as window, façade, balcony, corridor–serve as foundational visual building blocks for creating complex compositions, our framework aims to provide fundamental morphological elements that designers can combine to explore diverse robot forms.

To establish the design framework, we analyzed existing robots to identify fundamental building blocks of robot form that support creative design. We drew on the IEEE Robots Guide [28], a comprehensive database spanning diverse robot domains and commonly used in prior morphology research [*e.g.,* 14, 46]. From this collection, we included 170 out of 264 robots for the analysis after removing items outside our scope (*e.g.,* vehicles, modular robot kits, 3D printers). We started the analysis by printing images of the selected robots and physically laying them out to enable direct, hands-on comparison. We grouped robots based on silhouette and structural similarity, and this process yielded 18 groups. Next, for each group, we created abstract sketches that outlined element boundaries and basic forms using simple shapes (*e.g.,* representing a tabletop robot as a circle atop a rectangle), and annotated them with component-level functions and keywords (*e.g.,* "facial expression/gaze cues" for the circle, "stability" for the rectangle). We then clustered elements by functional and formal similarity, resulting in 13 categories, which were organized into four higher-level groups: perceptual elements that signal sensing and decision-making (*e.g.,* heads or sensor clusters); kinematic elements that support local motion or deformation; locomotion elements that enable movement through space (*e.g.,* wheels, legs, tracks); and structural elements that connect components and define overall body form. Through iterative refinement, we further divided the kinematic group into two subsets: articulation elements that produce movement or shape change, and end-effector elements typically located at the endpoint of moving parts to interact with objects or the environment.

From this process, we identified five fundamental elements of robot morphology: (1) **perception**: components where signal sensing and decision-making or embodies focus of attention; (2) **articulation**: components that create dynamic movements or large deformations; (3) **end effector**: parts that interact with objects or the environment; (4) **locomotion**: mechanisms that enable the robot to travel through space; (5) **structure**: components that organize and support other elements. By combining these five elements in various ways, existing robots can be described and analyzed, and designers can use them as building blocks to explore robot forms.

### 3.2 Morphology Exploration Blocks (MEB)

To operationalize the framework, we developed a tangible design tool called Morphology Exploration Blocks (MEB). Prior robot morphology frameworks are primarily organized through taxonomies, linguistic descriptors, or conceptual categories [*e.g.,* 16, 46], which limits opportunities for visual or physical exploration of robot form. To address this gap, we drew inspiration from the Graphic Design Play Book [12] and other card-based design tools [49] that translate abstract concepts into actionable, generative materials. Building on insights from creative toolkit research that emphasizes hands-on and reflective exploration [20], we envisioned a tangible medium that enables designers and researchers to actively manipulate, compare, and reflect on robot morphology.

The block designs were based on abstract sketches created during the framework development process, which captured recurring robot components in simplified form and common morphological features observed in existing robots. We abstracted the forms to remain neutral with respect to specific metaphors, functions, or contexts. We conducted two informal workshops to refine the shapes and verify that they were intuitive and not confusing for users. Figure 2 shows MEB, organized into five categories corresponding to Elements of Robot Morphology. Rather than creating a comprehensive library, we aimed to provide a representative set of blocks sufficient to support exploration. All blocks were 3D-printed with white PLA filament and embedded with magnets to allow easy connection and rearrangement during design activities.

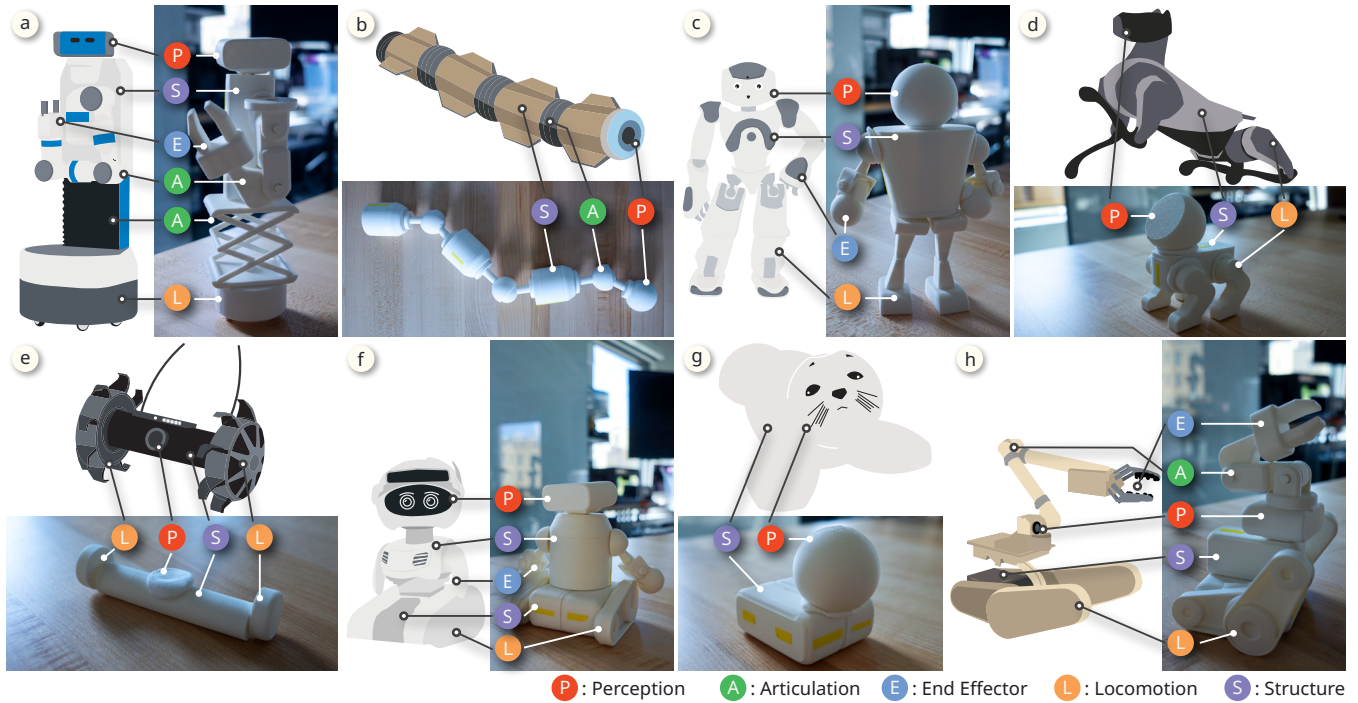

Figure 3: This figure highlights a subset of the annotated and reconstructed robots in the case study: (a) *Fetch*, (b) *ACM-R5H*, (c) *Nao*, (d) *CyberDog 2*, (e) *Throwbot*, (f) *Misty II*, (g) *Paro*, and (h) *Kobra*. (Robot illustrations are created by the authors, inspired by publicly available images [57–62, 64, 65].) The full catalog of results is available in our open science repository as supplementary material.

### 3.3 Case Study

We conducted a case study in which three paper authors, all HRI researchers, annotated and reconstructed robots using MEB. The study demonstrated how Elements of Robot Morphology framework and MEB serve as *analytical* tools, produced a catalog of block-based representations, and revealed limitations to refine both the framework and toolkit.

To capture a broad range of robot designs, we selected 24 robots from the IEEE Robots Guide [28], spanning eight locomotion types (aquatic, underbody wheels, wheels, tracks, bipedal, quadrupedal, aerial, and stationary) and four application contexts (industrial, research, service, and social). This selection ensured diversity in both morphology and use case. The five elements of our framework served as a coding scheme. After a brief orientation to align understanding, three researchers independently annotated all 24 robots. They then compared annotations, discussed disagreements, and refined interpretations of morphological features as needed. Following consensus, the researchers collaboratively used MEB to recreate each robot's morphology. Clay was used to supplement the blocks when a specific shape or connection could not be represented. Figure 3 presents examples from the case study. In the following, we summarize the key lessons from the case study. We believe these also serve as usage recommendations for our framework.

*Articulation & Locomotion: Conceptual Overlap.* We found that articulation and locomotion overlap conceptually. In this study, we define locomotion as the mechanisms that allow the entire robot to move or travel, whereas articulation involves the movement of specific parts of the robot, such as end effectors or a part of its body. This distinction helped keep the coding process consistent.

*Define the Threshold for "Dynamic" Movement.* Articulation in our framework is defined as "components that create dynamic movements or large deformations" in the robot's overall form. However, determining what counts as "dynamic" depends on the analysis context. For instance, we did not annotate simple head rotation as articulation because it represented a small, localized motion that a simple rotation can demonstrate. By contrast, we annotated Pepper's waist bend, as it significantly alters the robot's posture and expressiveness. It is therefore essential to clearly define the level of motion considered relevant before starting an analysis.

*Importance of Video Reference.* Video footage is critical for accurately annotating and recreating robots. Static images can be misleading, as interpretations of morphological elements change with viewpoint and motion. For example, the Picker Robot [63] appeared to lack a perceptual element in a static side view, but in motion, its end effector orientation suggested a face-like configuration, prompting anthropomorphic interpretation. Similarly, the Starship delivery robot [55] appeared featureless in bystander photos, yet from a user's perspective, its sensor array conveyed perception. These cases highlight the importance of considering spatial relationships and movement to avoid misinterpretation.

*Scale and Proportion Limitations.* Finally, while MEB was generally successful at representing the overall functions and relationships of robot components, scale and proportion emerged as notable limitations. For example, as shown in Figure 3(e), a block model may capture a robot's overall structure but fail to represent the proportion of wheels to the body.

## 4 Design Workshop

We conducted design workshops to understand how participants use MEB to generate and explore robot designs. We conducted both individual and group sessions (Figure 1(b)–(c)), observing how participants used the toolkit for individual ideation and collaborative creation. All research activities were reviewed and approved by the Ethical Review Board of the University of Wisconsin-Madison.[1]

### 4.1 Individual Design Workshop

*Participants.* We recruited 12 participants (five men, five women, two others) through a campus mailing list, ranging in age from 18 to 55 ($M = 29.00$, $SD = 10.21$), with one participant who did not report age. Participants represented three domains: four artists (animation, filmmaking, art education), four designers (architecture, UX, AR, accessibility), and four mechanical engineers (industrial design, manufacturing, soft robotics). Our intention in recruiting this population was to evaluate the expressive and technical range of MEB, and we expected these participants to explore a broad spectrum of possibilities. We expected them to be able to integrate our design tools into a creative process and provide a design rationale. All participants were geographically located in the United States and were fluent English speakers. The entire procedure took participants 60 minutes, and they received $20 USD.

*Procedure.* Each session began with a brief introduction to the study's goal: to explore and design robot forms using a toolkit. Participants were then presented with three design scenarios and asked to select one: (A) a patrol robot for crowded public spaces, (B) a delivery robot handling fragile objects, or (C) an educational assistant supporting classroom engagement. These scenarios were designed to reflect common and emerging robot applications, informed by iterative discussions within the research team and prior HRI work [30, 31]. After selecting a scenario, participants sketched initial concepts on a worksheet to explore the robot's appearance, functions, and interactions. They were then introduced to ten descriptive keywords drawn from the Godspeed Questionnaire [7], which served as anchors to structure the design process. These terms were chosen because they provide intuitive, validated descriptors widely used in HRI. Each participant selected one keyword, and the facilitator randomly assigned another, resulting in two keywords that guided the creation of two distinct robot designs based on the same concept. After the keywords were selected, the facilitator introduced MEB and its five categories. White air-dry clay was provided for sculpting custom parts or securing blocks. Participants then built two robot designs, one for each keyword, and finalized them by specifying physical properties (*e.g.,* size, material, texture), behaviors, and a name. Participants subsequently presented their designs, describing their concepts and design choices, while the facilitator asked follow-up questions about component functions, intent, and inspiration. If time permitted, participants could create a third robot using a different scenario. Each session concluded with a semi-structured reflection interview and a demographic survey.

### 4.2 Group Design Workshop

*Participants.* The group design workshop was held in an advanced sculpture course offered by the university's Art Department. Six participants, including a professor (sculptor), a technical staff (sculptor), and four students enrolling in the advanced sculpture course (ages 25-48, $M = 32.50$, $SD = 9.68$), took part, with two participants not completing the demographic survey. They were paired into three groups of two for a 45-minute session to collaboratively create robot forms using MEB. One pair left after the main activity, so the interview was conducted with the remaining two groups. Student participants received extra credit as an incentive.

*Procedure.* The group design workshop used the same design scenarios, keywords, and materials (*e.g.,* worksheets, MEB, and clay) as the individual workshop. The procedure was largely the same as the individual format, with several key differences. Each pair designed a single robot guided only by a self-selected keyword, without an additional randomly assigned keyword. After finalizing their design, groups participated in a brief share-out session, presenting their robot to the others. The workshop concluded with a group post-activity interview, followed by a demographic survey.

### 4.3 Data Collection and Analysis

For both the individual and group workshops, we collected audio recordings, video recordings, photographs of the final designs, and completed worksheets as data for analysis. Audio recordings were transcribed for qualitative analysis. Three researchers collaboratively annotated the photographs and transcriptions, identifying key themes, patterns, and design decisions.

## 5 Findings

In this section, participants in individual workshops are labeled P1–P12, and group pairs G1–G3. Designs in Figure 4 are indexed by scenario: patrol (A-1–A-5), delivery (B-1–B-17), and education (C-1–C-7). All quotes are italicized and in double quotation marks. Figure 5 summarizes MEB usage frequency.

### 5.1 How Participants Interacted with Blocks

*Exploration, Generation, Inspiration.* Participants used the blocks as hands-on tools to explore ideas through experimentation. Several described beginning without a clear concept and allowing form to emerge through play; *e.g.,* P2 noted, "*I didn't have an idea right away, so I just played with the parts until something came to me.*" Others found that physical manipulation helped translate abstract ideas into concrete designs. P1 explained that working with blocks prevented getting "*lost in nebulous concept generation*," while G2 contrasted the blocks with drawing, noting that they enabled them to "*imagine shapes I hadn't been intending on.*" Participants also described the blocks as creative prompts that expanded the design space. G2 reflected on the abundance of possibilities ("*I could use this... I could use that*"), and G1 emphasized that robot-inspired parts made designs feel more realistic and grounded, bringing them "*further along*" in the design process.

---

[1] Workshop materials, 3D printable block files, and case study results are available at https://osf.io/d3cwz/

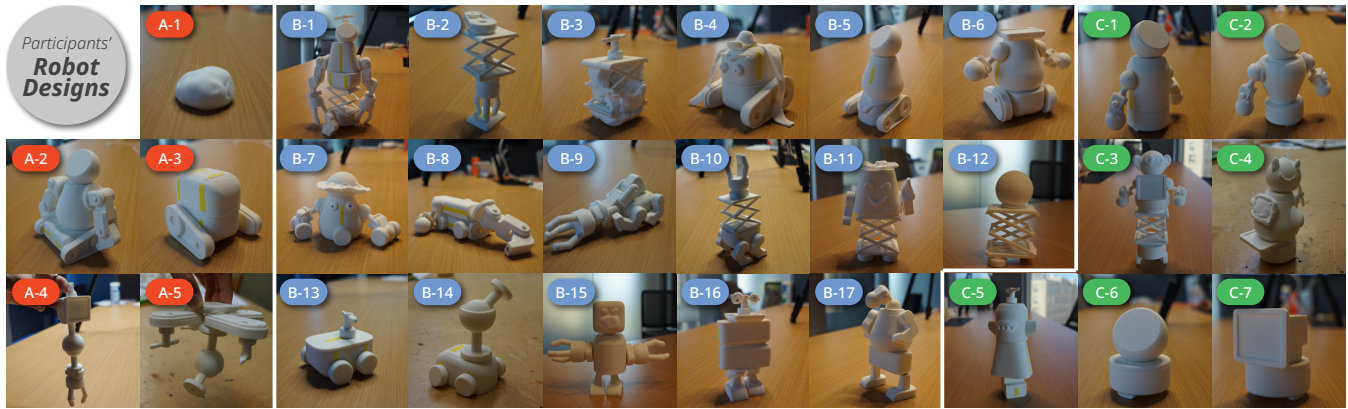

Figure 4: Participants' final robot designs. Designs are grouped by scenario: patrol (A-1–A-5), delivery (B-1–B-17), and education (C-1–C-7). In total, 29 unique robot designs emerged, illustrating the diversity of forms generated through open-ended exploration with MEB. A-5, B-14, and C-4 were created in the group workshop; all others were produced in individual workshops.

*Visual Communication and Collaboration.* The blocks provided a shared language for design discussions, enabling participants to communicate their ideas quickly and visually. Participants reported that it was faster and easier to mock up ideas with blocks than to describe them verbally or through drawings. G2 noted that "*we both would just grab one and swap it in and out [...] it came together without having to articulate it so precisely.*" Moreover, in the post-activity interview, most of the participants used their robot or block to communicate their idea to the facilitator.

*Challenges When Designing with* MEB. While MEB was perceived as generally helpful, participants identified several limitations. First, the strength and orientation of the magnets constrained how blocks could be connected. P6 and G1 noted that magnets were "*too weak*," while P9 mentioned that magnets that attached only in a single orientation limited exploration of different configurations. Second, participants encountered scale-mismatch issues. P7 found it confusing that blocks varied in scale, which made it challenging to design coherent robot forms. Finally, concerns about structural sturdiness disrupted the design process. For example, P8 changed their design mid-process when the underbody wheel block felt unstable.

## 5.2 Usage of Each Block

*Perception.* The tablet block appeared in 8 of the 29 designs. In three cases, it supported the robot's task by displaying information, such as a teacher's schedule (C-7) or the robot's internal state (C-3). Two designs used the block as an input surface, enabling users to type commands (B-6) or students to ask questions (C-4). In addition, the tablet block was repurposed in three designs, serving as a structural or symbolic element, including a drone wing (A-5), a connector between the body and face (B-15), or a landing pad for a flying robot (B-2). Circular shapes, including the truncated sphere, sphere, and hemisphere, appeared in 10 of the 29 designs and were commonly used to convey friendly, approachable, or inviting qualities (C-1—C-4). For example, P1 preferred a truncated sphere over a cube block, noting that "*the square is less approachable than the circle.*"

| Perception | | Locomotion | |
|---|---|---|---|
| Truncated Sphere | 6 | Wheels | 7 |
| Sphere | 3 | Underbody Wheels | 7 |
| Hemisphere | 1(1) | Tracks | 7(2) |
| Cube | 1 | Quadrupedal Legs | 5(2) |
| Flat Cuboid | 1 | Bipedal Legs | 3 |
| Tablet | 8(3) | Propellers | 7(2) |
| *Articulation* | | Fins | 2(2) |
| Two-link Manipulator | 4(2) | *Structure* | |
| Ball Joint | 3(1) | Right Angle | 2 |
| Jack Lift | 7(3) | Long Cuboid | 3 |
| *End Effector* | | Cuboid | 8(1) |
| High-Dexterity | 6 | Truncated Pyramid | 3(1) |
| Low-Dexterity | 0 | Thick Cylinder | 2(1) |
| Anthropomorphic | 6(2) | Truncated Cone | 0 |
| *Clay* | | Vase-shaped | 8(1) |
| Connection | 8 | Thin Cylinder | 0 |
| Add Detail | 8 | Thin Cuboid | 1 |
| New Material | 2 | | |

Figure 5: Usage frequency of MEB across 29 designs. Values indicate the total number of times each block was used; numbers in parentheses denote the subset of repurposed uses (*e.g.*, 8(3) indicates eight total uses, including three repurposed).

*Articulation.* Articulation blocks appeared in 14 of the 29 designs. Among them, the jack lift block was used in seven designs. It commonly enabled height adjustment, allowing delivery robots to reach objects (B-1, B-10, B-11) or education robots to align with children's eye level, avoiding "*looking up or down*" and conveying an equal relationship (C-3). In other cases (B-2, B-3, B-12), the same vertical motion was repurposed as an "*expand and contract*" mechanism to enclose and accommodate delivery items. The two-link manipulator appeared in four of the 14 designs. In two cases, it served as an alternative to a human-like arm; however, participants selected it specifically to evoke "*scary*" (A-2) or "*anxious*" (B-9) impressions. The ball joint appeared in three of the 14 designs. It was used functionally to control an end effector (A-4) or move a surveillance camera (A-5), and expressively as a form of "*contraction*" to convey "*playfulness*" (B-14). Although used infrequently, several participants expressed interest in the ball joint; for example, P1 noted, "*I couldn't figure out a way to incorporate that, but this [the ball joint] was interesting to me.*"

*End Effectors.* End effector blocks appeared in 12 of the 29 designs. The high-dexterity end effector was used in six designs and applied as intended, enabling actions such as "*grasp and drop*" (A-4) or handling "*fragile objects with complex fingers*" (B-1). Although the low-dexterity end effector did not appear in any final designs, participants frequently compared it with the high-dexterity version during ideation. The human-like arm was used in six designs. While often contrasted with the two-link manipulator, it was selected when participants aimed to support expressive gestures or pointing in human–robot interaction contexts (C-1—C-3), rather than object manipulation. P12 expressed interest in more human-like arms resembling an "*artist's mannequin.*" However, limitations in arm length were sometimes perceived negatively; *e.g.,* P7 described the arm as conveying "*unintelligence*" due to being "*not long enough.*" Several participants also repurposed the human-like arm in unconventional ways, such as using it to support stable locomotion with wheels (B-7) or as a robot head (B-17).

*Locomotion.* Locomotion blocks appeared in 28 of the 29 designs, serving as the primary means of movement, though participants' reasons for selecting specific locomotion types varied. Some choices were driven by expressive intent: *e.g.,* P4 selected a quadrupedal form to convey anxiety, inspired by experiences with the CyberDog robot, while also exploring continuous tracks to achieve a softer, more approachable appearance. Functional considerations such as stability and efficiency also shaped locomotion decisions. P8 emphasized stability by combining continuous tracks (B-4) with multiple wheels (B-6), while P6 highlighted energy efficiency, noting that wheels are commonly perceived as the most efficient mode of robotic movement (B-13). Design context further influenced locomotion choices. Education robots frequently employed underbody wheels to enable smooth navigation and rotation in classroom environments (C-1). P7 explored adaptive locomotion by designing a robot capable of switching between quadrupedal movement on uneven terrain and wheels on flat surfaces (B-1), while opting for a simpler tracked design in a hotel service scenario (B-7). Participants also creatively repurposed locomotion blocks as anthropomorphic or expressive elements. For instance, quadrupedals were used as robot arms (B-11, B-17), fins as short arms (C-4), and a propeller to structure a robot's face (B-16) or evoke a "*wind-up toy*" (C-4).

*Structure.* Structure blocks appeared in 23 of the 29 designs and primarily served to organize other elements, while also enabling a range of creative reinterpretations. Seven designs envisioned structures as functional spaces such as fountains, charging stations, or book storage (C-1, C-7), and frequently used them as containers for delivery items (B-4—B-6, B-10, B-13). Two designs employed structure blocks as protective casings for internal components (B-1) or as housings for sensors, cameras, and defensive tools (A-3). Structure blocks were also used to form anthropomorphic heads with features such as eyes or mouths (B-4, B-6, B-11). The block shape further influenced the perceived robot impression. For example, G1 oriented a curved block upward to create a "*less threatening*" appearance (B-14), while the drop-shaped block was commonly associated with friendly, interactive designs (A-2, C-3, C-4), often evoking Baymax from Big Hero 6. In contrast, P11 inverted the same shape to convey "*authority,*" such as that associated with teachers (C-2). Finally, combining multiple structure blocks enabled diverse body forms and expressive meanings. Vertical stacking was used to suggest "*intelligence*" (B-11), while other participants constructed chassis-like bases (A-2, A-3, B-4) or distinctive aesthetics, such as a "*1950s retro toy*"–inspired design (P12). Overall, these combinations highlight the versatility of structure blocks in shaping both functional and expressive aspects of robot design.

*Clay.* Clay was used in 15 of the 29 designs and was valued as a flexible connector, allowing participants to explore configurations beyond the constraints of the magnetic blocks. As P8 noted, "*the addition of the clay really rounded out the kit very well.*" Beyond structural connections, 8 designs used clay to add expressive and decorative details. Participants sculpted features such as eyes, hair, hats, and facial elements to convey distinct personalities, ranging from intentionally "*unfriendly*" (B-4) to "*organic*" (B-6) or playful, child-like appearances (C-4, B-17). P2 imagined clay as a material with futuristic functions: *i.e.,* "smart clay" (A-1) or shock-absorbing components (B-3), expanding the role of clay beyond decoration and assembly into generating design ideas.

### 5.3 Design Outcomes and Patterns

Across the three scenarios, the delivery robot scenario was selected most frequently (eight individuals and one group), followed by the education robot (three individuals and one group) and the patrol robot (two individuals and one group). The popularity of the delivery scenario likely reflects participants' familiarity with sidewalk delivery robots, while those with teaching experience were drawn to the education scenario. Overall, participants tended to choose scenarios that felt familiar and easy to imagine.

Many designs adopted anthropomorphic structures with distinct heads, torsos, limbs, and locomotion (A-2, B-1, B-5, B-6, B-12, B-15, B-17, C-1–C-3), reflecting human-like archetypes. Some participants further exaggerated or stylized these forms to emphasize character or personality (B-4, B-7, B-11, B-16, C-4, C-5). Other designs closely resembled existing robots, drawing on participants' prior knowledge (A-3, A-5, B-7, B-9, B-10, B-13, B-15), while several adopted futuristic or science-fiction aesthetics (A-4, B-2, B-3, B-8). A smaller subset explored compact, minimalist forms (A-1, C-6, C-7).

## 6 Discussion
### 6.1 Framework Supports for Form Exploration

Our framework and blocks supported a shift from conceptual ideation to tangible exploration. By bridging imagination and concrete design, MEB enabled designers to embody ideas through hands-on iteration. This was particularly valuable for participants without a clear initial concept or those seeking to move beyond familiar mental images, as physical play encouraged experimentation and unexpected discoveries [17]. Together, these observations highlight MEB's role as an early-phase design aid that helps externalize abstract ideas and iteratively refine them. The diversity of blocks fostered creativity by balancing inspiration and constraint, enabling designs that were both imaginative and functionally grounded. In this sense, the framework offers a structured lens for reasoning about robot form, while MEB provides a tangible means of exploration; together, they support design-space navigation through activities such as comparing forms, examining articulation strategies, and reflecting on morphological trade-offs. Beyond design exploration, the framework and blocks supported reflection and analysis. Participants' designs revealed underlying assumptions about form and meaning (*e.g.,* associating greater height with intelligence), while reconstructing robots in the case study highlighted the role of spatial reasoning and three-dimensional relationships. Examining these patterns offers researchers insight into how designers connect perception, metaphor, and morphology. Finally, MEB functioned as a shared design language that facilitated communication among designers and between designers and researchers. By externalizing morphological decisions in a tangible and interpretable form, the toolkit supports a range of early-phase activities, including planning, concept generation, and early prototyping, and can aid cross-disciplinary communication with engineers, clients, or other stakeholders. In educational contexts, MEB also holds potential as a tool for teaching shape language and foundational design principles, supporting STEAM learning.

### 6.2 Metaphor-morphology Connection

We observed that metaphors and metaphorical reasoning are pervasive in the design processes and outcomes. Metaphors offer templates for robot form design (*e.g.,* human-like form) and serve as anchors that help designers generate and build up ideas [3, 35, 36]. For example, we observed extensive use of *shape language*—the use of form to communicate the robot's personality, mood, or intent. Some participants explicitly mentioned their experiences and uses of the *shape language*, while others appeared to use it through metaphors; these metaphors seemed to help them intuitively apply *shape language*, linking familiar narratives to tangible form exploration [14]. This pattern suggests that while MEB can support designers in morphologically realizing their conceptual, metaphorical ideas, metaphors can also serve as accessible entry points into morphological exploration. To facilitate this creative transition, in future work, our design framework can be complemented by structured approaches to metaphorical design [*e.g.,* 3, 14] to help designers' morphological exploration. However, they must be balanced with considerations of real-world functionality. Robots ultimately operate in physical environments, and designs cannot rely solely on symbolic or narrative meaning. Thus, our framework and blocks offer a complementary lens for breaking down robots into elemental components, thereby complementing metaphorical thinking.

### 6.3 Limitations & Future Work

***Limitations of Our Framework.*** A key limitation of Elements of Robot Morphology is the lack of visual and tangible qualities, such as color, texture, or material properties of robots, that play an important role in shaping robot character and expressivity, particularly in contemporary HRI research that leverages material properties, *e.g.,* soft actuators or shape-changing skins to convey emotion, intent, or personality [25, 29, 32]. Furthermore, our framework is based on the robot databases we referenced. Robotics is a rapidly advancing field, with emerging forms such as origami robots, mechanical metamaterials, reconfigurable systems, and other unconventional embodiments. Casting a wider net would enable more comprehensive future frameworks. We envision our framework as an evolving structure, continually revised as robotics and material-science innovations introduce new possibilities for robot form.

***Limitations of MEB.*** A limitation of MEB is that, unlike existing tangible design materials such as LEGO Serious Play [19] that allow designers to construct an unlimited range of forms, pre-determined blocks may limit form exploration. We note that, these materials emphasize additive construction, making it difficult to deliberately subtract, reconfigure, or compare alternative morphological choices in a systematic way. In contrast, our goal was to scaffold the design process by creating design anchors that can enable quick, hands-on visualization, exploration, and comparison at early stages of robot design. That said, future work can extend MEB in several ways. First, future versions can integrate smaller components, such as facial features and sensors, and flat or thin pieces. Participants frequently used clay to sculpt elements like smiling faces and repurposed the tablet and wheel blocks as makeshift joints or structural elements. Second, MEB can be extended to consider elements such as color and material by incorporating complementary tools such as color charts and swatch cards [*e.g.,* 11]. Finally, future versions of MEB must convey scale and proportion, as the uniform dimensions of the blocks prevented consideration of relative size relationships between components, which may constrain design exploration.

***Study Design Limitations*** The design scenarios used in the workshops did not capture the full range of real-world robot applications. For example, we did not include stationary or tabletop robots, which are common in social and educational settings, nor scenarios in workplaces or caregiving contexts. Expanding the scenario set would enable future studies to evaluate how the framework and blocks perform across a broader spectrum of robot types and deployment settings. Participant recruitment also presents a limitation of this study. To evaluate the expressive and technical range of MEB, we intentionally recruited participants with creative backgrounds. While appropriate for assessing the toolkit's generative potential, this focus limits insight into how MEB might be used by stakeholders without design experience. Prior HRI research shows that co-design with domain stakeholders can produce more contextually grounded and actionable insights [*e.g.,* 1, 24, 41]. Future studies should explore scenario-specific workshops with relevant stakeholder groups, such as caregivers, educators, or end-users, to better understand how MEB supports robot design beyond design experts. Finally, this study relied on designers' prior knowledge when selecting and combining blocks, providing no guidance (*e.g.,* "underbody wheels enable smooth travel") so participants interpreted blocks based on their own experience. Future work could integrate design principles–like those in graphic design–into the framework to offer richer guidance on how forms combine and what meanings and experiences they evoke.

Elements of Robot Morphology: Supporting Designers in Robot Form Exploration HRI '26, March 16–19, 2026, Edinburgh, Scotland, UK

## Acknowledgements
We would like to thank the participants and the People and Robots Lab for generously sharing their time, creativity, and perspectives in this study.